\documentclass{article}

\usepackage[preprint]{nips_2018}

\usepackage[utf8]{inputenc} 
\usepackage[T1]{fontenc}    
\usepackage{hyperref}       
\usepackage{url}            
\usepackage{booktabs}       
\usepackage{amsfonts}       
\usepackage{nicefrac}       
\usepackage{microtype}      
\usepackage{amsmath}
\usepackage{graphicx}
\usepackage{subcaption}
\usepackage{epstopdf}

\title{Human Action Generation with Generative Adversarial Networks}

\author{
  Mohammad Ahangar Kiasari, Dennis Singh Moirangthem, Minho Lee\\
  School of Electronics Engineering\\
  Kyungpook National University\\
  80 Daehakro, Bukgu, Daegu - 41566, South Korea \\
  \texttt{\{ahangar100, mdennissingh, mholee\}@gmail.com} \\
}

\begin{document}

\maketitle

\begin{abstract}
Inspired by the recent advances in generative models, we introduce a human action generation model in order to generate a consecutive sequence of human motions to formulate novel actions. We propose a framework of an autoencoder and a generative adversarial network (GAN) to produce multiple and consecutive human actions conditioned on the initial state and the given class label. The proposed model is trained in an end-to-end fashion, where the autoencoder is jointly trained with the GAN. The model is trained on the NTU RGB+D dataset and we show that the proposed model can generate different styles of actions. Moreover, the model can successfully generate a sequence of novel actions given different action labels as conditions. The conventional human action prediction and generation models lack those features, which are essential for practical applications.
\end{abstract}

\section{Introduction}

Generating novel sequences of human motion to form an action has been a challenging problem. Human action classification and prediction have been studied in detail but there have been few works exploring generation of novel actions. With the advent of powerful generative models such as generative adversarial networks (GAN)~\citep{goodfellow2014generative}, novel data generation has become a possibility. However, in human action generation, simple generation of a random sequence of motions may not suffice the requirements and usability. We need a proper control of the action labels for the generated actions and also the style of the motion sequences in a given action class.

Recent studies have explored the possibility of human action prediction given a number of prior motion sequences ~\citep{butepage2017deep,cvpr2017human,barsoum2017hp}.~\citet{butepage2017deep} proposed a sparse autoencoder model to predict human motions in an unsupervised manner.~\citet{cvpr2017human} proposed a prediction model based on sequence to sequence(Seq2seq) model to predict short-term human actions. On the other hand,~\citet{barsoum2017hp} introduced a Seq2seq model trained with an adversarial cost to predict the future skeleton frames given a sequence of prior frames.

The conventional human motion prediction models described above have several limitations as they are proposed to just predict the future frames given a series of inputs. These models have limited ability to generate novel actions. Moreover, such models have no control, such as the class of action or initial position, over the predicted motions as the generation is conditioned over the input frames. The recently proposed prediction model~\citep{barsoum2017hp} that utilizes GAN in the model to enhance the generation capability does not have the capability to generate a desired class of action. Furthermore, the generated future motion sequences are highly dependent on the given prior poses and the effect of random $z$ on the generated action style is minimal. In order to address these issues, we need a generative model that can handle generation of desired action classes  with different styles.

The aim of this paper is to explore a new algorithm to generate multiple classes of human actions as well as consecutive sequences of human actions. Indeed, most of the movement information is stored in the human skeleton sequences. Therefore, generating corresponding human skeleton plays an important role in the human motion generation task. This fact inspired us to introduce a model that is able to generate a novel sequence of human skeleton poses with the help of a generative model.

In this work, we introduce a new framework to generate novel human actions with the help of GAN. We propose a model, which combines an autoencoder and a conditional GAN~\citep{mirza2014conditional}, in order to generate a sequence of human actions. The autoencoder and the conditional GAN are trained simultaneously in an end-to-end manner. The model generates human actions based on a given action class and an initial start position. We can change the random vector $z$ to generate different styles of actions. The proposed model can also generate a sequence of human actions without any continuity issues. We train our model using the NTU RGB-D dataset~\citep{Shahroudy_2016_CVPR}, which is collected using Microsoft Kinect, and we show that the model can successfully generate a variety and combinations of multiple actions.

The major contributions of this paper are as follows:
\begin{itemize}
\item We introduce a semi-supervised model based on GAN to generate novel human actions.
\item The proposed model takes into account the class of the action to be generated as well as the initial state.
\item Our model has the capability to generate multiple styles of a single action class by changing the random vector $z$.
\item It can also generate a consecutive sequence of different actions with the help of the initial position and different class labels during generation. This allows the model to generate a series of action sequences with seamless transition.
\end{itemize}

\section{Related works}
\label{gen_inst}

Recurrent neural networks(RNNs) such as long short-term memory(LSTM)~\citep{hochreiter1997long} or gated recurrent units(GRU)~\citep{bahdanau2014neural} have been at the forefront of human motion prediction. However, such deep neural networks based model are primarily deterministic. There have been attempts to modify RNN encoder-decoder frameworks to work as a combination of deterministic and probabilistic human motion prediction models~\citep{iccv2015recurrent}. In recent advances,~\citet{cvpr2016structural} and~\citet{cvpr2017human} have introduced several structures and frameworks with deep RNNs to produce state-of-the-art results in human motion prediction. On the other hand~\citet{butepage2017deep} proposed a sparse autoencoder model to predict human motions in an unsupervised manner without using RNNs. Even though these models are good in human motion prediction, generation of novel human actions have been a challenge with such models.

Generative Models have received a significant boost in performance and applicability with the advent of Generative Adversarial Networks(GAN)~\citep{goodfellow2014generative,mirza2014conditional,arjovsky2017wasserstein,li2017mmd,liu2017unsupervised}. Recent studies have incorporated GAN in RNN sequence-to-sequence (Seq2seq)~\citep{sutskever2014sequence} architectures to improve motion prediction. Researchers in~\citep{barsoum2017hp} developed a model called HP-GAN to achieve the probability of the future sequences conditioned on the given incomplete input sequence. In order to predict, the model basically employed a Seq2seq framework with mapping random $z$ vector on the encoder output part of the Seq2seq model. However, the model is not able to generate new actions based on different class labels. Moreover, generating multiple sequences without the prior human poses is not considered in this study.

One of the recent works~\citep{cai2017deep} introduced human motion prediction using GAN trying to close the gap between prediction and generation. The model contains a two-step generation pipeline. The human pose generation part consists of two generative models, which are trained separately to predict a sequence of human skeleton. The generative models have the random input $z$ along with two conditions called $z_0$ and $c$. The $z_0$ and $c$ present initial pose of the generated sequence and the class label, respectively. However, this part of the model is not an end-to-end algorithm. moreover, the paper lacks sufficient exploration regarding the effect of $z_0$ and $c$ in the generation process.

Despite the use of RNN and GAN, the current models lack the complete features of a true generative model for human actions. We develop an end-to-end model that can generate different human actions with different styles. Our generative model is based only on GAN and we can generate particular classes of human motions conditioned on the given class label and initial pose, and generate different styles in each class with a variable random $z$. The amount of control and flexibility introduced in our model make it more usable in real applications.

\section{Methodology}
\label{headings}
\begin{figure}
\centering
\includegraphics[width=0.5 \columnwidth]{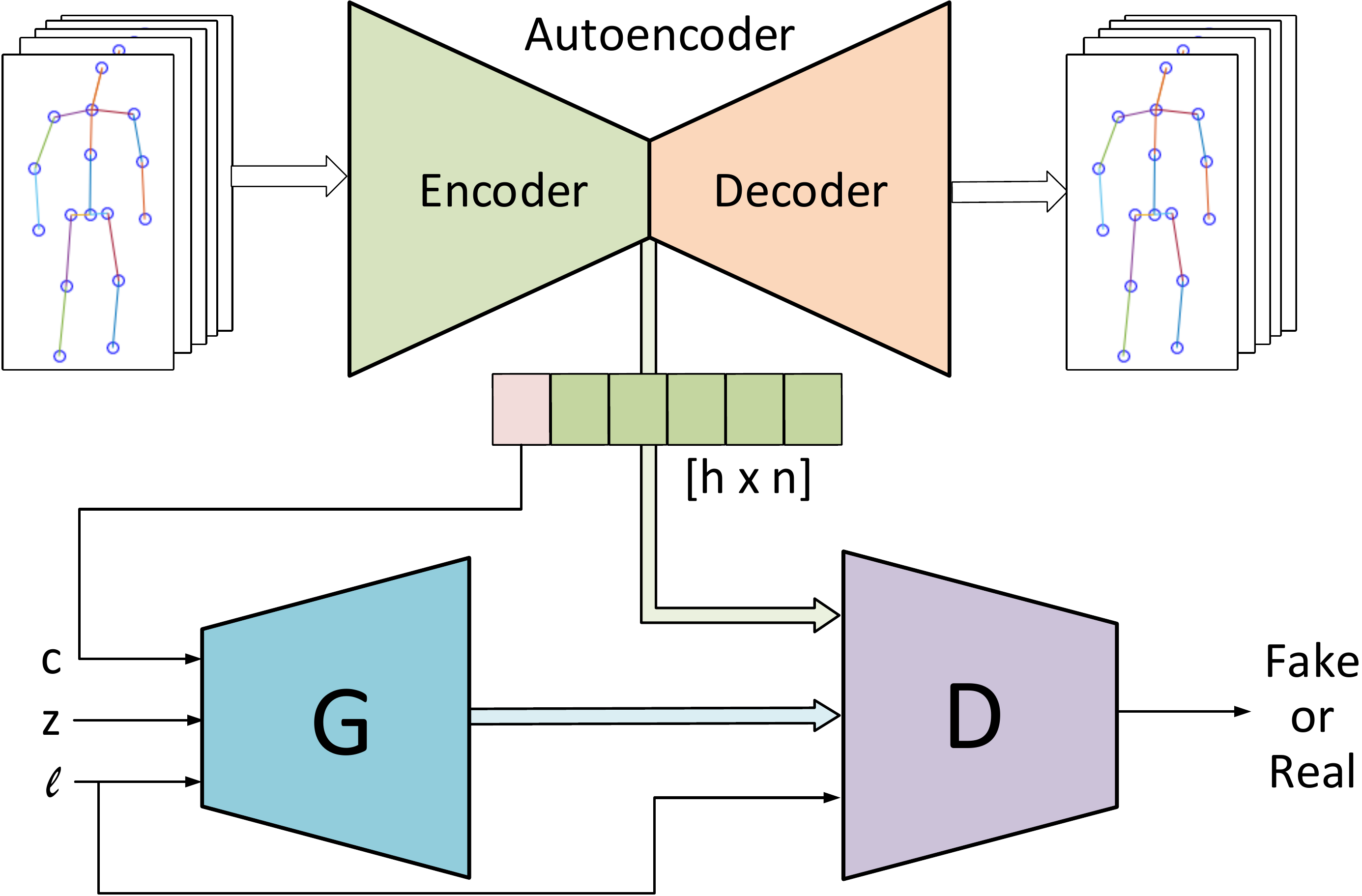}
\caption{The proposed model consists of an autoencoder and a conditional GAN that can take multiple conditions to generate multiple classes of human actions with different styles.}
\label{fig1}
\end{figure}

In this section, we describe the proposed model, which consists of an autoencoder and a GAN as shown in Fig.~\ref{fig1}. All the components of the model are trained simultaneously end-to-end. The trained model is then configured to the generation phase in order to produce the human motions.

\subsection{Generative Adversarial Networks (GAN)}

Generative adversarial networks(GANs)~\citep{goodfellow2014generative} present a push-pull game between a generator and a discriminator. The training objective of the discriminative model is to determine whether the data are from fake data generated by the generative model or real training data. For the generative model, its objective is to generate realistic data, which is similar to the true training data and the discriminative model can't distinguish. For the standard generative adversarial networks, we train the discriminative model $D(\cdot)$ to maximize the probability of giving the correct labels to both the samples from the generative model and training examples. We simultaneously train the generative model $G(\cdot)$ to minimize the estimated probability of being true by the discriminator. The objective function is:
\begin{equation}
\min_G \max_D V(D,G) = E_{x\sim p_{data}(x)}[\log D(x)] + E_{z\sim p_z(z)}[\log(1-D(G(z)))]
\label{eq:gan}
\end{equation}
where $z$ denotes the random input data with a uniform or normal distribution $p_{z}(z)$, and $G(\cdot)$ denotes the generator. $D(\cdot)$ indicates the discriminator and $x$ is the training data with empirical distribution $p_{data}(x)$. 


\subsubsection{Conditional Generative Adversarial Networks}

In the conventional GAN, there is no control on the mode of the generated data. Therefore,~\citep{mirza2014conditional} proposed the conditional generative adversarial networks to address this problem. In the conditional GAN, both the input random $z$ to the generator and the input of the discriminative model are concatenated with a label vector. The label vector contains the mode of the data such as class labels. Recently, several papers successfully applied conditional GAN on different purposes~\citep{chen2016infogan,odena2016conditional,makhzani2015adversarial}. In this paper, we also utilize the conditional GAN to introduce control on the class label and initial pose of the generated action sequences. 

\subsection{Proposed model}

We employ an adversarial generative model to address human action generation problem. The goal of the model is to generate sequences of human poses with random input $z$ conditioned by an initial pose. The initial pose condition is introduced in the model in order to enable the model to generate consecutive multiple human actions without continuity problems. To this purpose, we develop a framework of a conditional generative adversarial networks and an autoencoder. 

The input and output of the autoencoder are vector representation of human-pose frames. The reconstruction loss function of the autoencoder is an $L_2$ norm. The latent variables in the bottleneck of the autoencoder, $h$, represents the low-dimensional space of the original vector of each frame of human pose. Therefore, with a sequence of frames as the input of the autoencoder, we can get a sequence of low dimensional vector, $h_{seq}$, at the latent space. The dimension of the $h_{seq}$ is $n\times N$, where $n$ is the dimension of $h$ and $N$ is the length of each sequence. We utilize this low-dimensional representation of the original skeleton as the real data in the discriminator. Fig.~\ref{fig1} demonstrates the architecture of the proposed model. 

The generative adversarial model includes a generator and a discriminator. The input of the generator is a concatenation of three vectors. The first vector is the $z$ vector drawn from a random normal distribution. The second vector $c$, represents the initial state of the human pose. The third vector $l$ is the action class label. The class-label is a one-hot vector corresponding to the number of actions defined in the dataset. We concatenate the generator output, $\hat{h}_{seq}$, with the class label and the initial pose and feed it into the discriminator as fake data. We optimize the parameters of the discriminator and generator using min-max optimization given in Eq. (\ref{eq:loss_gan}).

\begin{equation}
\begin{aligned}
\min_G \max_D \text{E}_{\mathbf{x} \sim p_{seq} } [\text{log} D(\mathbf{x|c,l})] + \text{E}_{\mathbf{z} \sim p_z} [\text{log} (1 - D(G(\mathbf{\mathbf{z|c,l}}))]
\label{eq:loss_gan}
\end{aligned}
\end{equation}

where $G(\cdot)$ and $D(\cdot)$ are the Generative and Discriminative models, $p_{seq}$  and $p_z$ represents the distribution of real sequences and the random data distribution, respectively. As a result, the generator learns to imitate human action sequences with respect to the given initial posture state $c$ and action label $l$. The aim of employing input $z$ is to generate a variety of action styles with a fixed $c$ and $l$. In addition to the generator and discriminator losses, we add another loss to keep the consistency between adjacent frames in each generated sequence as shown in Eq. (\ref{eq:loss_cons})~\citep{barsoum2017hp}.  

\begin{equation}
\begin{aligned}
L_{cons} = \frac{1}{M(N-1)} \sum_{i=1}^{M} \sum_{j=1}^{N-1}(s_{ij}-s_{i(j+1)})^2
\label{eq:loss_cons}
\end{aligned}
\end{equation}

where, $s_{ij}$ shows the $j^{th}$ frames in $i^{th}$ generated sequence. $M$ is the batch size and $N$ is the length of $h_{seq}$. Since, $N$ can be a large value, it can dramatically increase the computation cost. Hence, in order to compute the $L_{cons}$ in each iteration, instead of the full sequence, we randomly select a part of each sequence with starting point $S$ and length $L$. Eq. (\ref{eq:loss_cons1}) shows the new consistency loss.

\begin{equation}
\begin{aligned}
L_{cons} = \frac{1}{M(L-1)} \sum_{i=1}^{M} \sum_{j=S}^{L-1}(s_{ij}-s_{i(j+1)})^2
\label{eq:loss_cons1}
\end{aligned}
\end{equation}

Eq. (\ref{eq:total_loss}) shows the total objective function in the generative and discriminative model.

\begin{equation}
\begin{aligned}
\min_G \max_D V(D,G) = \text{E}_{\mathbf{x} \sim p_{seq}} [\text{log} D(\mathbf{x|c,l})] + \text{E}_{\mathbf{z} \sim p_z} [\text{log} (1 - D(G(\mathbf{\mathbf{z|c,l}}))] + \lambda L_{cons}
\label{eq:total_loss}
\end{aligned}
\end{equation}

where, the $\lambda$ is a value between 0 to 1. We found that a very small $\lambda$ results in losing consistency between adjacent frames in a sequence. Reversely, by choosing a very large value, the generator cannot learn different patterns of sequences. As a result, the generated sequences are almost constant even with different random $z$. We set 0.01 to $\lambda$ in all our experiments.

\begin{figure}
\centering
\includegraphics[width=0.5 \columnwidth]{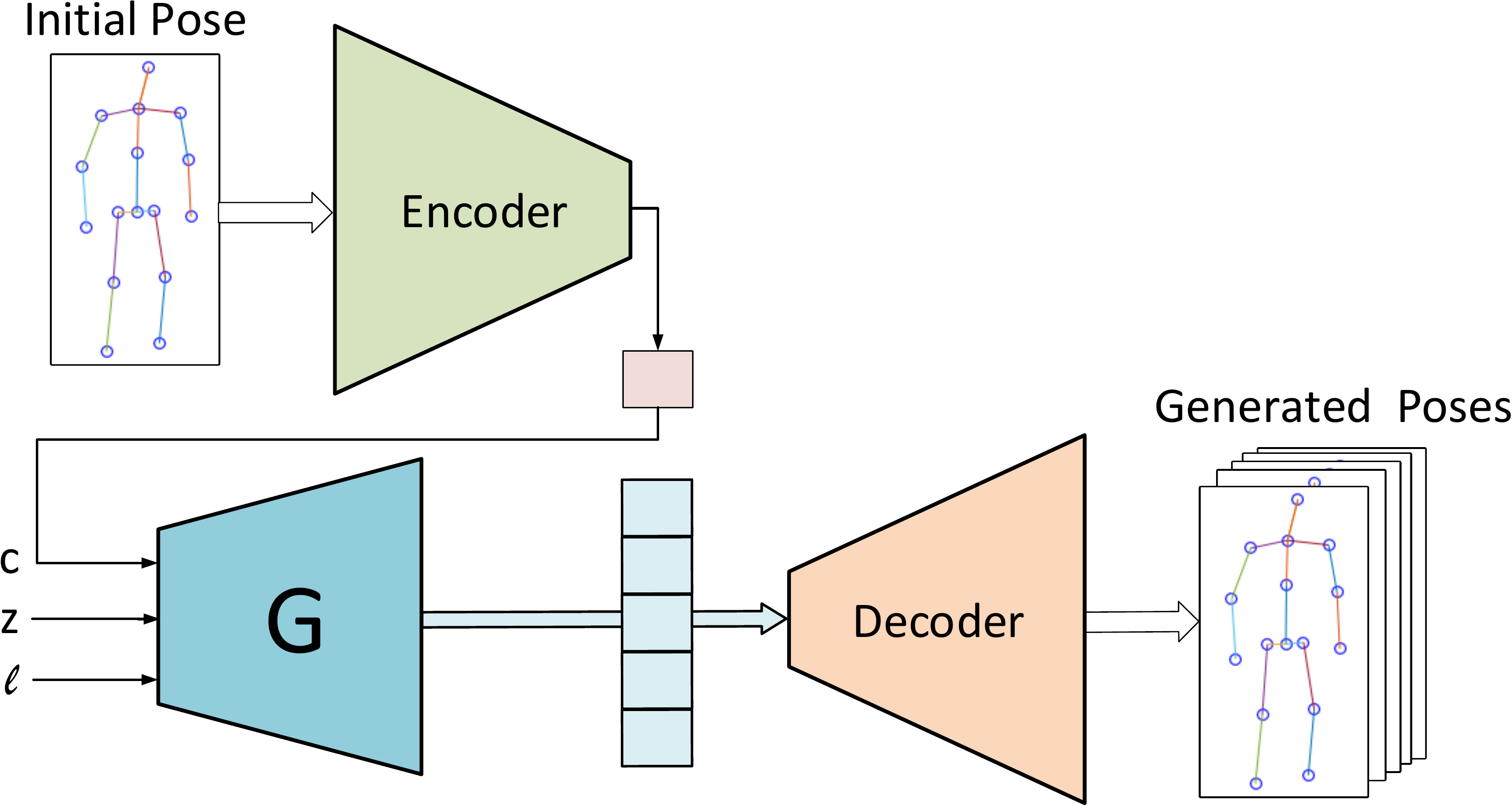} 
\caption{The model in generation phase utilizes the encoder and the decoder to produce the generated actions using the generator $G(\cdot)$.}
\label{fig:generate}
\end{figure}

In the generation phase of our model shown in Fig.~\ref{fig:generate}, we use the trained encoder to generate the initial condition $c$ from a single given frame. The initial condition $c$, random $z$ and the label of the desired class $l$ are fed into the trained generator $G(\cdot)$ to generate the output $\hat{h}_{seq}$. Then this $\hat{h}_{seq}$ output is passed on to the decoder to construct the sequence of generated human poses. 

\section{Experiments and results}
We evaluate our model based on the benchmark NTU RGB+D dataset~\citep{Shahroudy_2016_CVPR}. The NTU RGB+D action recognition dataset consists of $56,880$ action samples containing RGB videos, depth map sequences, 3D skeletal data, and infrared videos for each sample. This dataset was captured by 3 Microsoft Kinect V2 cameras concurrently. We utilize the 3D skeletal data, which contains the three dimensional locations of 25 major body joints, at each frame. The dataset contains $60$ different action classes including daily, mutual, and health-related actions collected using $40$ different actors. 

All parameters of the model are trained using the Adam optimizer(ADAM)~\citep{kingma2014adam}. We set the learning rate as $0.00001$. The ReLU activation~\citep{nair2010rectified} is used in the network with the exception of the output layers of autoencoder, generator and discriminator, where we applied $tanh$, $linear$ and $sigmoid$ activation functions, respectfully. We applied a normal distribution with zero mean and unit variance to generate random inputs $z$. The model is trained with a mini-batch size of $64$.

\begin{figure}[!ht]
\centering
\includegraphics[width=0.5 \columnwidth]{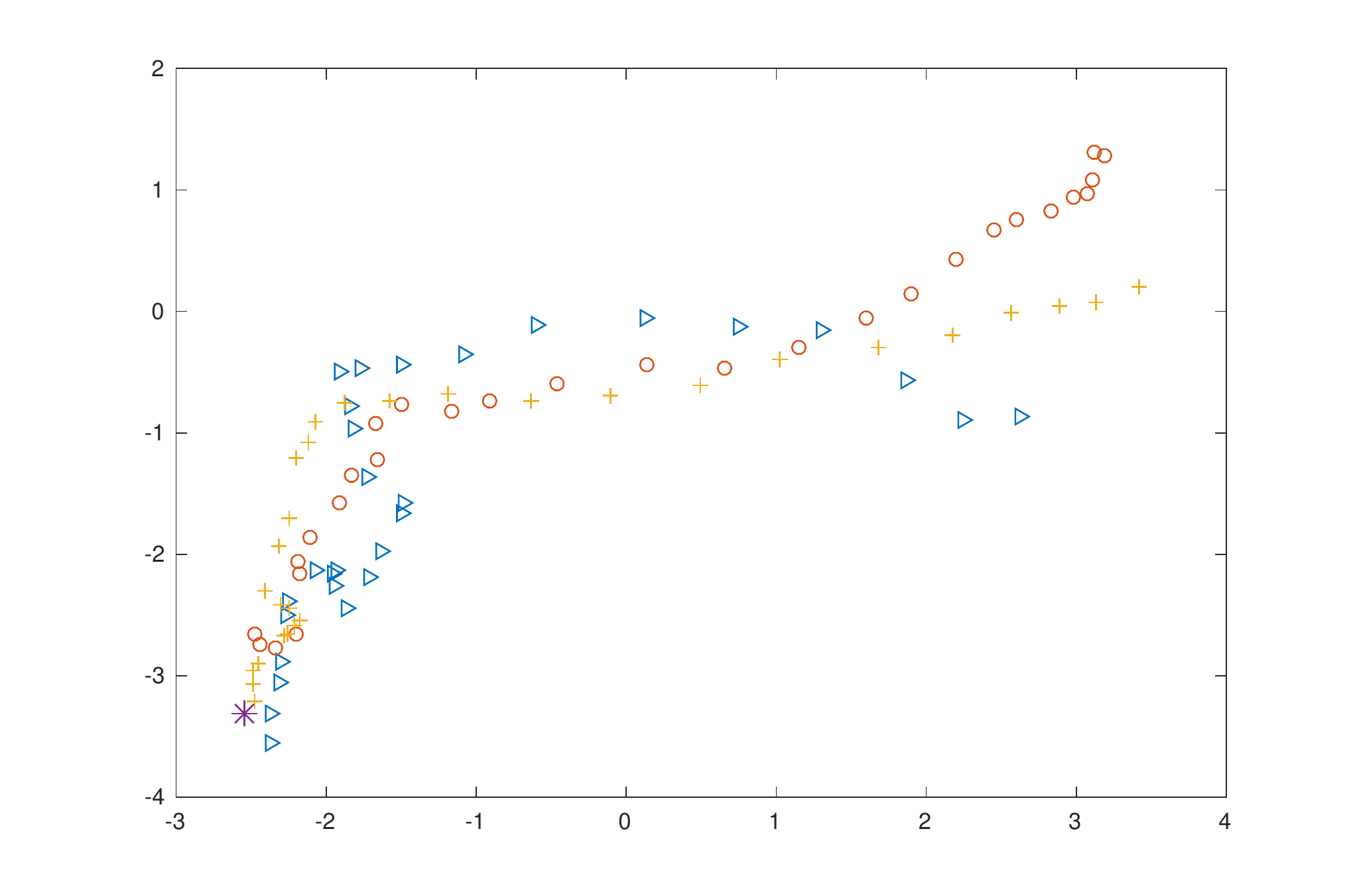} 
\caption{Generated sequences mapped into 2-dimensional space with three different random $z$ and a fixed initial pose $c$. The star marker shows the given initial pose. The generated sequences are specified by different colors and markers.}
\label{routes}
\end{figure}

The entire model is trained end-to-end where the autoencoder and the GAN are jointly optimized on their corresponding losses. Once the model is trained, generation of new actions are performed using the generate phase of the model shown in Fig.~\ref{fig:generate}. In a generative model like the one proposed in this work, the effect of the random $z$ should be clearly demonstrated in the generated results. By using $z$ effectively, we will be able to generate different action styles that may not be included in the training data. Therefore, in order to evaluate the impact of the $z$ space in the sequence generation task, we generate three sequences with three different random vector $z$, drawn from Gaussian distribution, and a fixed initial condition $c$. Fig.~\ref{routes} illustrates the effect of applying different $z$ to generate various motion sequences in 2-dimensional space. The star marker in the figure indicates the given initial position $c$. According to these results, all of the three generated sequences started relatively close to the given initial pose, but at the later point of time during generation, the sequence changes. Hence, these results indicate that the proposed model is able to generate sequences with different styles using multiple $z$.

\begin{figure}[!ht]
\centering
\includegraphics[width=1 \columnwidth]{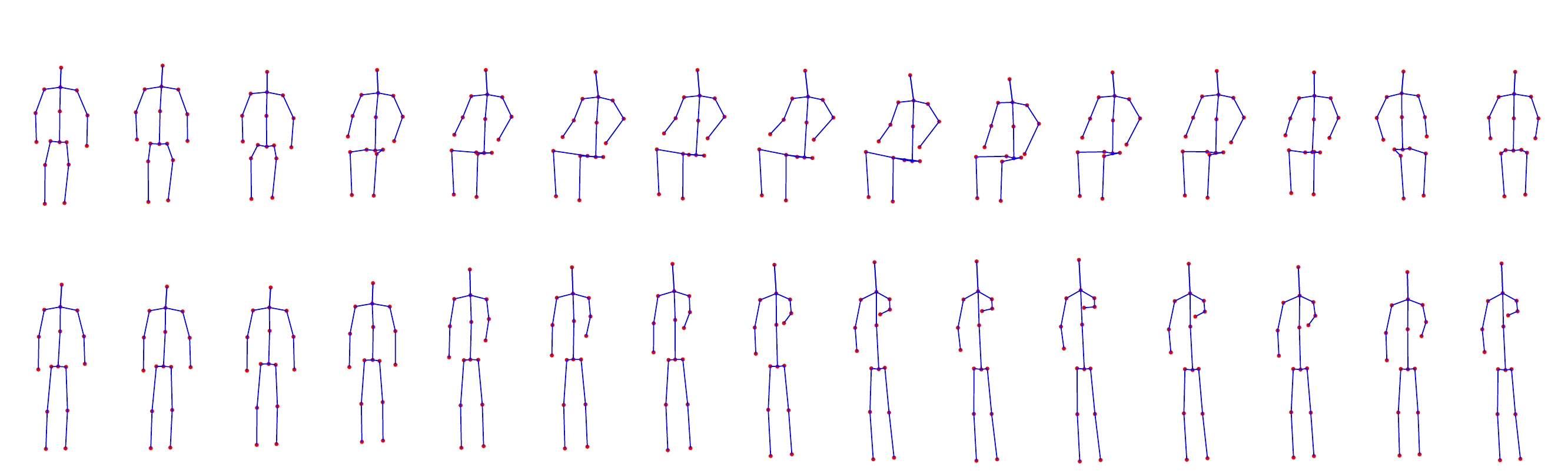} 
\caption{Generated actions of Sitting-down and throwing with identical initial pose $c$. The results show the impact of the changing class label $l$ in the generative model with the same initial pose condition.}
\label{sit_throw}
\end{figure}

Further experiments are conducted to show the effect of the class label $l$ on the generator output. Fig.~\ref{sit_throw} demonstrates the generated sequences with the same initial pose but with different class labels. The first and the second rows in Fig.~\ref{sit_throw} show the sitting down and throwing actions, respectively. The generated sequences not only follow the different corresponding actions but also start relatively near to the given initial pose.

\begin{figure}[!ht]
\vspace{-0.1in}
\centering
\begin{subfigure}{1\textwidth}
\centering
\includegraphics[width=1 \columnwidth,height=2cm]{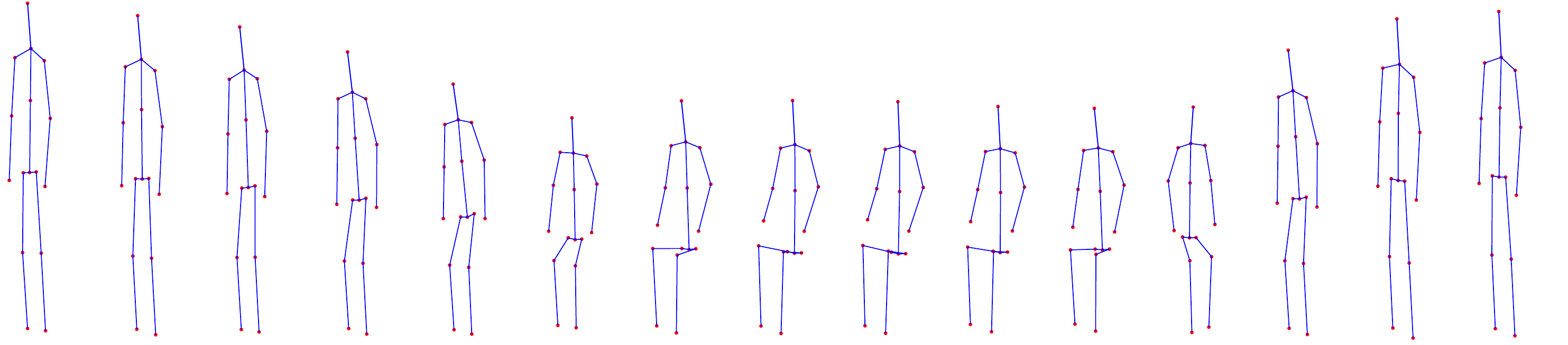}
\caption{}
\end{subfigure}
\centering
\begin{subfigure}{1\textwidth}
\centering
\includegraphics[width=1 \columnwidth,height=2cm]{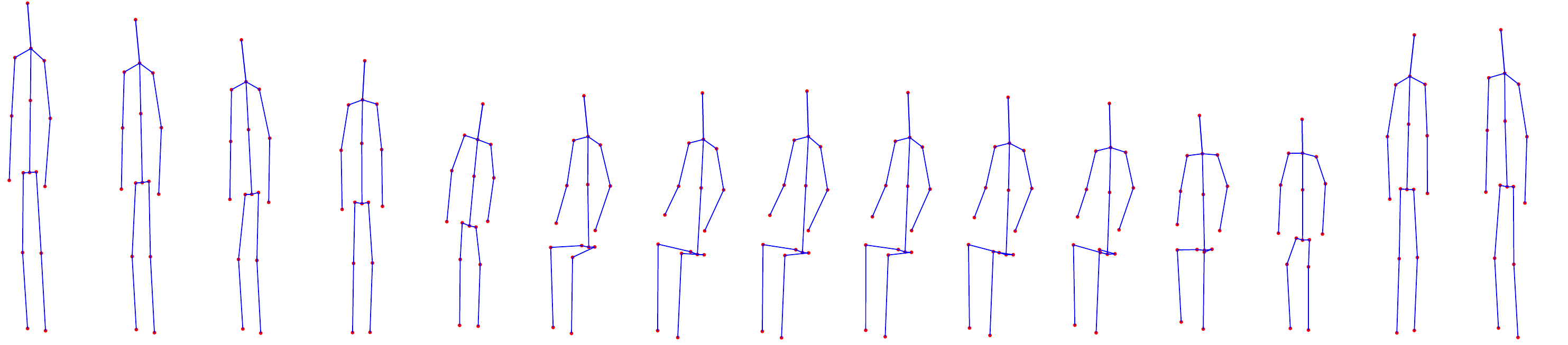}
\caption{}
\end{subfigure}
\caption{Generated sitting-down followed by standing-up sequences with two different random $z$. The initial pose conditions are identical in both (a) and (b) sequences. Even though (a) and (b) show the same sequence of actions, the generated results demonstrate the different styles using different $z$.}
\label{stand_sit}
\end{figure}

\begin{figure}[!ht]
\vspace{-0.1in}
\centering
\begin{subfigure}{1\textwidth}
\centering
\includegraphics[width=1 \columnwidth,height=2cm]{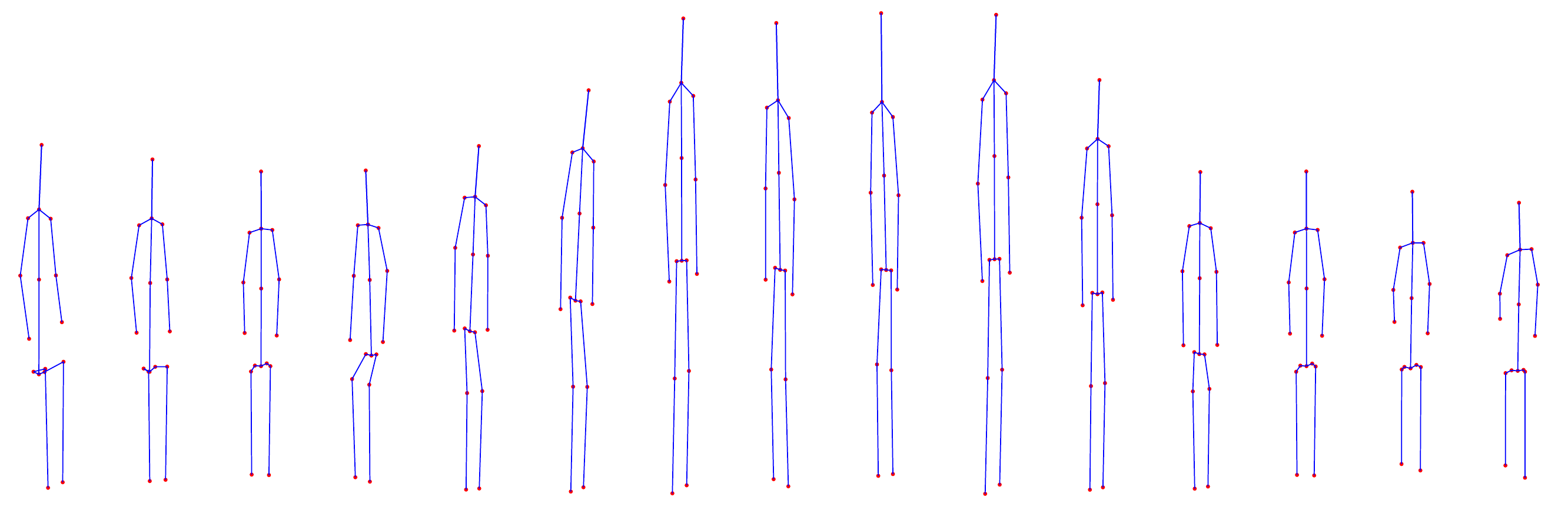}
\caption{}
\end{subfigure}
\centering
\begin{subfigure}{1\textwidth}
\centering
\includegraphics[width=1 \columnwidth,height=2cm]{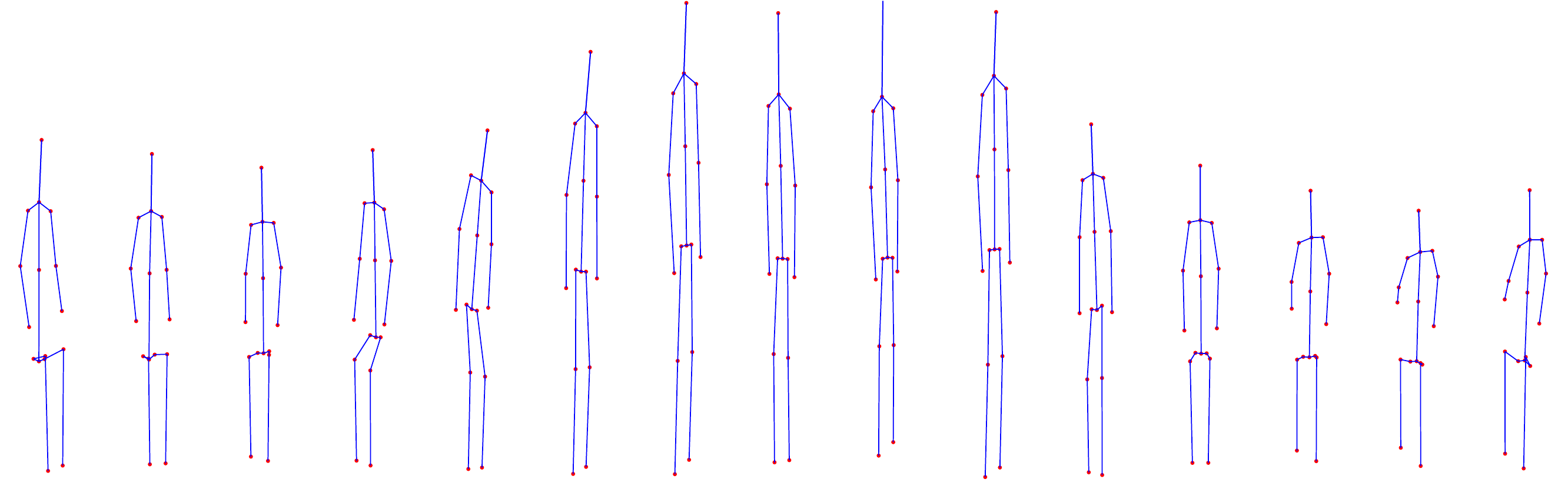}
\caption{}
\end{subfigure}
\caption{Generated standing-up followed by sitting-down sequences with two different random $z$. The initial pose conditions are identical in both (a) and (b) sequences.}
\label{sit_stand}
\end{figure}

In order to generate two consecutive actions, after generating the first action sequence, we have to reinitialize both $c$ and $l$ with the last frame of the first generated sequence and the new class label, respectively. Fig.~\ref{stand_sit} shows the standing-up action followed by sitting-down action with identical initial poses and two different random $z$. Similarly, Fig.~\ref{sit_stand} illustrates the results of sitting-down and standing-up with a random $z$.

\section{Discussion and future work}
The results of the proposed model illustrates several improvements over the existing human action prediction models~\citep{barsoum2017hp,cai2017deep} that uses generative models like GAN. The proposed model is able to generate a novel sequence of human motions without any prior sequence of input frames. The model also utilizes only a GAN for the generative part whereas other models utilizes multiple GANs~\citep{cai2017deep} as well as RNNs~\citep{barsoum2017hp,cai2017deep}. Another important feature of the proposed framework is the ability to generate multiple actions in a seamless consecutive manner. This major feature is lacking in the existing models.

\begin{figure}[!ht]
\centering
\includegraphics[trim={3cm 0 0 0},clip,width=1.1 \columnwidth]{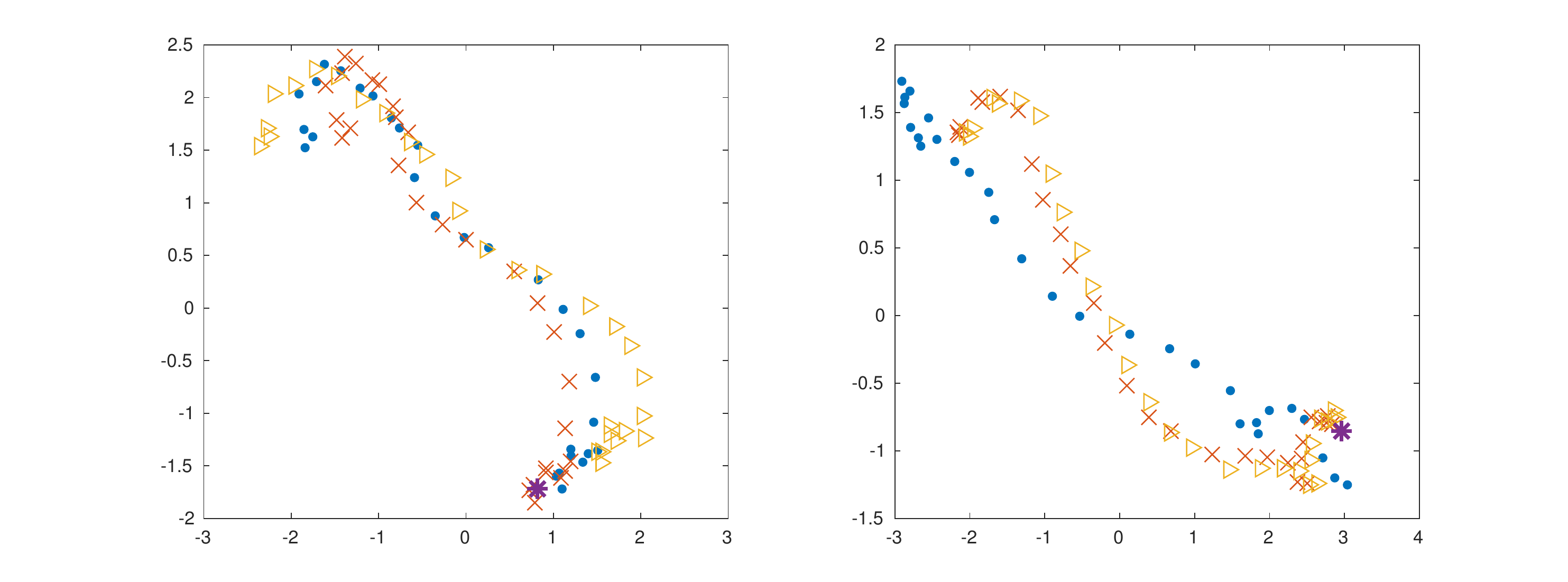} 
\caption{Generated sequences mapped into 2-dimensional space with three different random $z$ and two different initial poses. The star markers show the given initial poses. Each generated sequences are specified by different colors and markers.}
\label{multi_routes}
\end{figure}

The proposed model's ability to generate different styles of actions using different random $z$ has been well illustrated in the results. Moreover, the result shown in Fig.~\ref{routes} clearly demonstrate the generative capability where the generated actions differ considerably from each other with different random $z$ given the same initial conditions and class. Additionally, we demonstrated the generated sequences in 2-dimensional space with different initial poses $c$ and random $z$ in Fig.~\ref{multi_routes}. The results of both Fig.~\ref{multi_routes}(a) and Fig.~\ref{multi_routes}(b) indicate that the generated sequences started from different locations corresponding to the given initial poses. This result is also important to show the effect of the initial condition $c$ in the generation process where different $c$ results in different start positions of the generated sequence. As far as we know, these kind of analysis and results have not been demonstrated in prior works.

In the future, we will try to generate a blend of multiple actions, which are a combination of two or more action classes. We also plan to generate videos with the help of skeleton to image transformation models~\citep{yan2017skeleton}.

\section{Conclusion}

We introduced a new human action generation model that can generate a sequence of novel human actions. Unlike existing human action prediction models that use RNNs, we proposed a framework of an autoencoder and a conditional GAN trained end-to-end to produce multiple and consecutive human actions. The model trained on the NTU RGB-D dataset showed that it can generate different styles of actions. Furthermore, the proposed model was able to generate a set of consecutive actions with seamless transition given different action labels as conditions. We also showed the effect of random $z$ in the generated results where different $z$ produced different styles of actions.

\bibliographystyle{apalike}
\bibliography{mybibfile}
\end{document}